\documentclass[10pt,a4paper,conference,composocconf]{IEEEtran}

\usepackage{cite}

\ifCLASSINFOpdf
   \usepackage[pdftex]{graphicx}
  \graphicspath{{img/}}
  \DeclareGraphicsExtensions{.pdf,.jpeg,.png}
\else
\fi

\usepackage{amsmath}
\usepackage{array}

\usepackage{url}

\usepackage[utf8]{inputenc}

\usepackage{amssymb} %
\usepackage{mathtools}
\usepackage{bm}
\usepackage{xspace}
\usepackage{microtype}
\usepackage{multicol}
\usepackage{booktabs}
\usepackage{nth}

\usepackage[table]{xcolor}

\usepackage{footnote}
\usepackage{afterpage}

\usepackage{multirow}
\usepackage{tikz}
\usepackage{pgfplots}
\usepgfplotslibrary{fillbetween}
\pgfplotsset{compat=newest}
\usetikzlibrary{patterns,
pgfplots.groupplots,fit,calc,positioning,shapes,spy}

\usepackage{tabularx}
\usepackage{booktabs}

\usepackage{varioref}
\usepackage[pagebackref=true,breaklinks=true,letterpaper=true,colorlinks,bookmarks=false,citecolor=green!80!black,linkcolor=red!80!black,urlcolor=blue]{hyperref}
\usepackage{cleveref}
\usepackage{subcaption}
\usepackage{csquotes}
\usepackage{siunitx}
\let\labelindent\relax %
\usepackage[inline]{enumitem}
\setlist*[enumerate]{label=(\arabic*)}

\newcommand{\onedot}{.\xspace}
\newcommand{\etal}[1]{#1~et~al\onedot}

\newcommand{\cf}{cf\onedot}
\newcommand{\ie}{i.\,e.,\xspace}

\crefname{section}{Sec.}{Sections}
\crefname{figure}{Fig.}{Figure}
\crefname{table}{Tab.}{Table}
\crefname{equation}{Equ.}{Equation}

\newcommand{\map}{mAP\xspace}

\definecolor{faublue}{RGB}{0,51,102}
\definecolor{darkgreen}{rgb}{0,0.6,0}
\definecolor{bblue}{HTML}{4F81BD}
\definecolor{rred}{HTML}{C0504D}
\definecolor{ggreen}{HTML}{9BBB59}
\definecolor{ppurple}{HTML}{9F4C7C}

\NewDocumentCommand{\rot}{O{60} O{1em} m}{\makebox[#2][l]{\rotatebox{#1}{#3}}}%
\NewDocumentCommand{\rotn}{O{90} O{1em} m}{\makebox[#2][l]{\rotatebox{#1}{#3}}}%
\NewDocumentCommand{\rotninety}{O{90} O{1em} m}{\makebox[#2][l]{\rotatebox{#1}{#3}}}%

\begin{document}
\title{ICFHR 2020 Competition on Image Retrieval for Historical Handwritten Fragments}

\author{\IEEEauthorblockN{Mathias Seuret\IEEEauthorrefmark{1}, Anguelos Nicolaou\IEEEauthorrefmark{1},
Dominique Stutzmann\IEEEauthorrefmark{2}, Andreas Maier\IEEEauthorrefmark{1}, Vincent Christlein\IEEEauthorrefmark{1}}
\IEEEauthorblockA{
\IEEEauthorrefmark{1} Pattern Recognition Lab, Friedrich-Alexander-Universität Erlangen-Nürnberg,
91058 Erlangen, Germany\\
\{firstname.lastname\}@fau.de
}
\IEEEauthorblockA{
\IEEEauthorrefmark{2} Institut de Recherche et d'Histoire des Textes (Centre National de la Recherche Scientifique - UPR841), France\\
dominique.stutzmann@irht.cnrs.fr
}}

\maketitle

\begin{abstract}
This competition succeeds upon a line of competitions for writer and style analysis of historical document images.
In particular, we investigate the performance of large-scale retrieval of historical document fragments in terms of style and writer identification.
The analysis of historic fragments is a difficult challenge commonly solved by trained humanists. 
In comparison to previous competitions, we make the results more meaningful by addressing the issue of sample granularity and moving from writer to page fragment retrieval.
The two approaches, style and author identification, provide information on what kind of information each method makes better use of and indirectly contribute to the interpretability of the participating method.
Therefore, we created a large dataset consisting of more than 120\,000 fragments.
Although the most teams submitted methods based on convolutional neural networks, the winning entry achieves an mAP below 40\,\%.
\end{abstract}


\IEEEpeerreviewmaketitle

\section{Introduction} \label{sec:introduction}
Finding image fragments is of great importance for cultural heritage projects. It applies not only for finding coalescent fragments (so-called joints) as in the well-known Cairo Genizah collection of fragments, but also, if we focus on the handwriting, this challenge is applicable to the very many small pieces of text that are included in otherwise well-preserved codices, in the form of marginal annotations or interlinear glosses. From the point of view of the humanities scholar, these annotations can be considered palaeographically as fragments, with a piece of text and undefined margins. 
Therefore, the outcome may have direct impact on our knowledge of the past. 
Especially, in the age of mass digitization, a successful retrieval can assist humanists in their daily work.

The goal is to (a) find all fragments corresponding to a specific writer using a fragment from this writer as query and (b) find all fragments of corresponding to the same image.

This competition shares similarities with the work of \etal{Wolf}~\cite{Wolf11}, who proposed a system to identify join candidates of the Cairo Genizah collection. 
Therefore, the authors used writer identification as a part of a larger framework to find these candidates in approximately \num{350000} fragments of mainly Jewish texts.
The competition is also similar to previous ICDAR competitions on writer identification/retrieval~\cite{Fiel17ICDAR,Christlein19Comp} and also shares some similarities to word spotting. 
Similarly to word spotting, the challenge is to conduct an efficient image retrieval on small image patches.

The last competitions on historical image retrieval~\cite{Christlein19Comp} consisted of about \num{21000} document images from about \num{10500} writers.
In this competition, we increased the number of samples significantly (especially for training deep neural networks). Therefore, we employed a semi-automatic procedure that allowed us to generate in total more than \num{120000} fragments of about \num{9800} writers from about \num{20000} document images.

The paper is organized as follows. In \cref{sec:data}, we give details about the data generation process, its sources and the dataset splits. The submitted methods are explained in \cref{sec:methods}. \Cref{sec:evaluation} shows the results of the competition as well as additional experiments on data subsets. The paper is concluded in \cref{sec:conclusion}.

\section{Data}\label{sec:data}
The dataset was created semi-automatically and is publicly available.\footnote{\url{https://doi.org/10.5281/zenodo.3893807}}

\subsection{Fragment Generation}
Our goal is to generate fragments with random shapes, and non-straight edges.
We have developed a free, open-source fragmentation system,\footnote{\url{https://github.com/seuretm/diamond-square-fragmentation}} which is heavily based on the diamond-square algorithm~\cite{fournier1982computer}.
This algorithm is typically used to produce height maps in video games.

\begin{figure*}
\centering
\includegraphics[height=4cm]{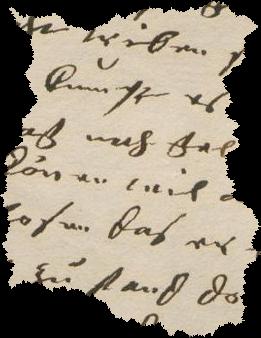}
\hfill
\includegraphics[height=4cm]{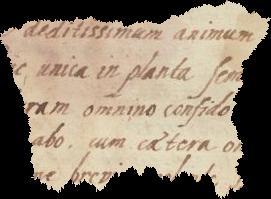}
\hfill
\includegraphics[height=4cm]{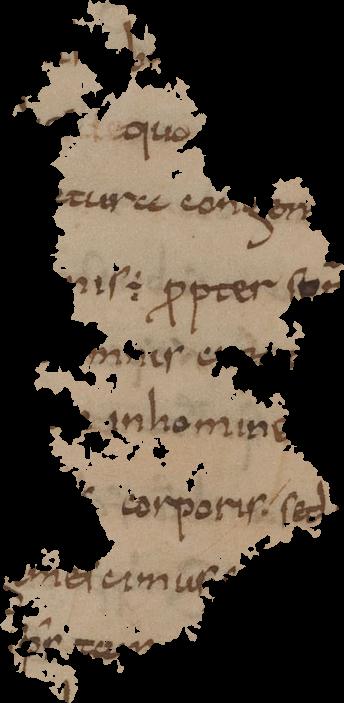}
\hfill
\includegraphics[height=4cm]{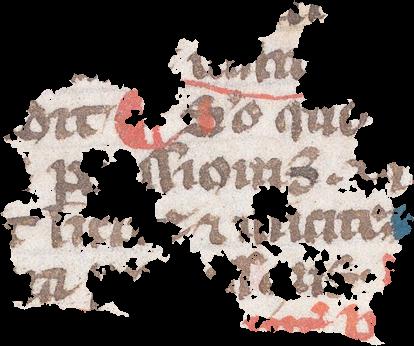}
\caption{Illustration of the two main kinds of fragments. The two left ones have rectangle-like shapes, while the ones on the right have totally random shapes. These fragments were randomly selected from the test set.}
\label{fig:frag-examples}
\end{figure*}

We produced two kinds of fragments.
The first kind has completely random shapes, without other restriction than being in one piece.
They can have holes.
The second kind has a more rectangular shape, and is produced by cutting the considered document using horizontal and vertical non-linear polygonal chains.
These chains have the restriction that they must always move forward along one axis.
Examples are given in \cref{fig:frag-examples}.

Our fragment generation was done in three steps.
First, we extracted fragments with random shapes from all images.
Second, we removed the ones which we did not want to keep (e.g., small ones as we can expect them not to contain enough information).
And third, we re-fragmented the pages with the lower amount of fragments  (after deleting their remaining fragments) using the method which produces rectangle-like fragments.

The two fragmentation methods are detailed below.

\subsubsection{Random-Shape Fragments}
For every document to fragment, we produce a 768$\times$768 image using the diamond-square algorithm.
Then, we resize with linear interpolation this image such that its new dimensions match the larger text area dimension.
We then crop its center to get the same shape as the text area.

\begin{figure*}
\centering
\begin{subfigure}{.2\textwidth}
  \centering
  \includegraphics[height=4cm]{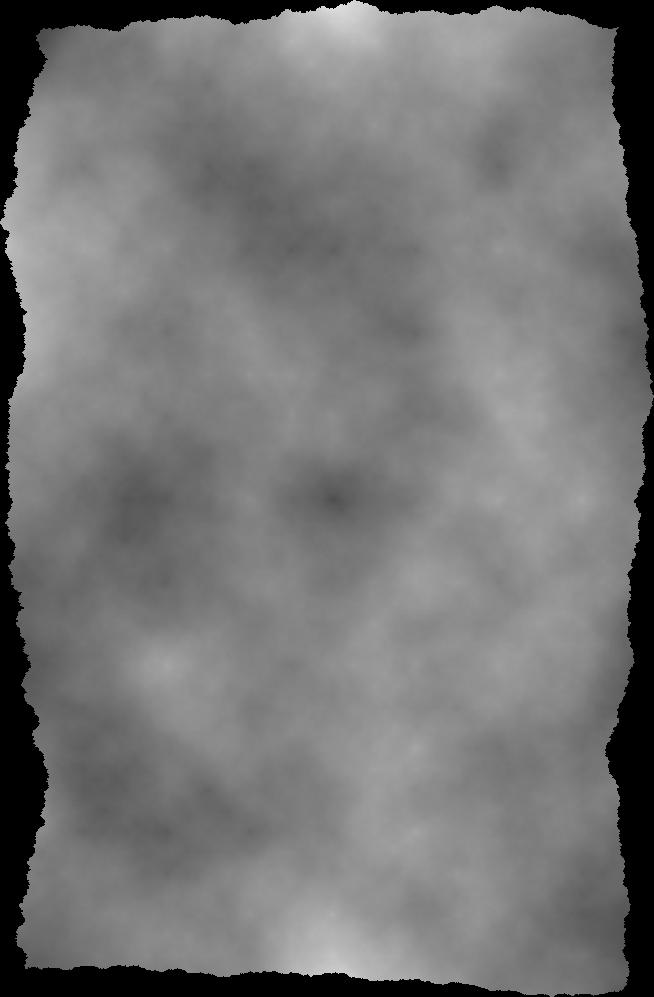}
  \caption{Diamond-square}
\end{subfigure}%
\begin{subfigure}{.2\textwidth}
  \centering
  \includegraphics[height=4cm]{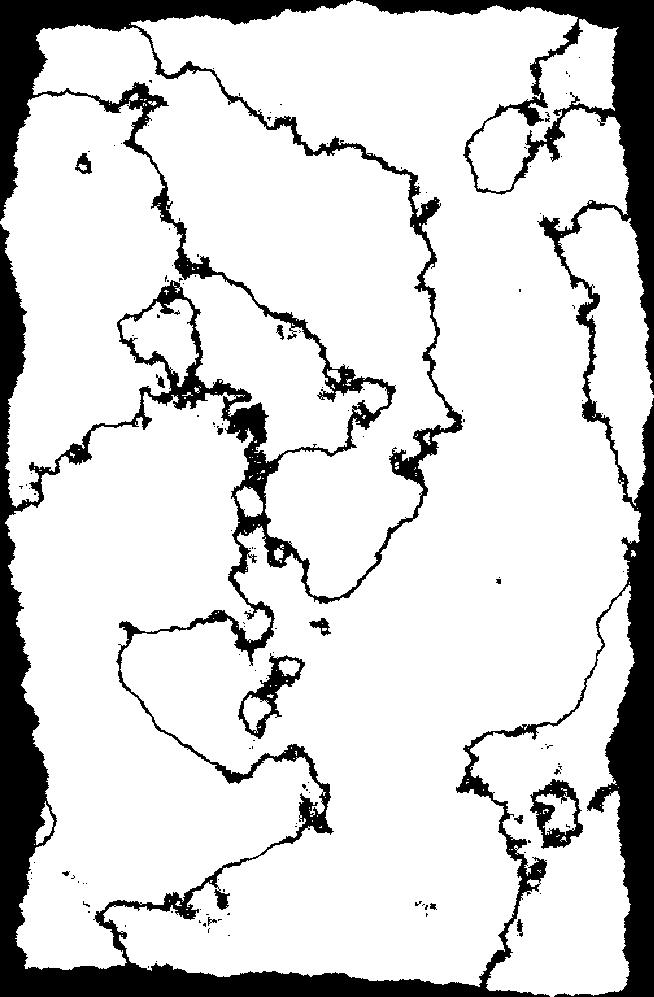}
  \caption{Margin = 1}
\end{subfigure}%
\begin{subfigure}{.2\textwidth}
  \centering
  \includegraphics[height=4cm]{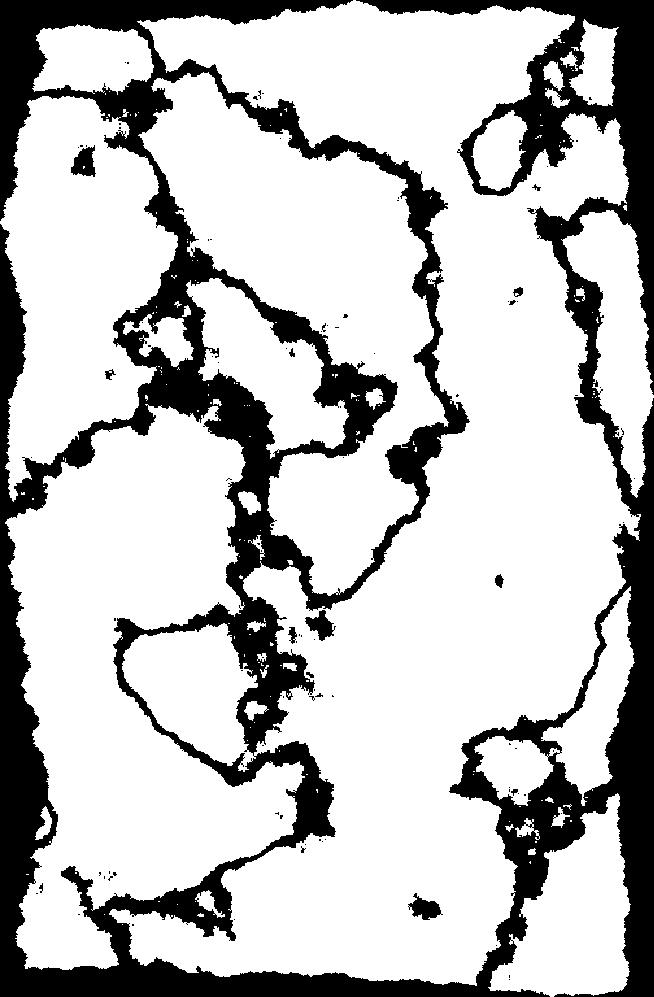}
  \caption{Margin = 3}
\end{subfigure}%
\begin{subfigure}{.2\textwidth}
  \centering
  \includegraphics[height=4cm]{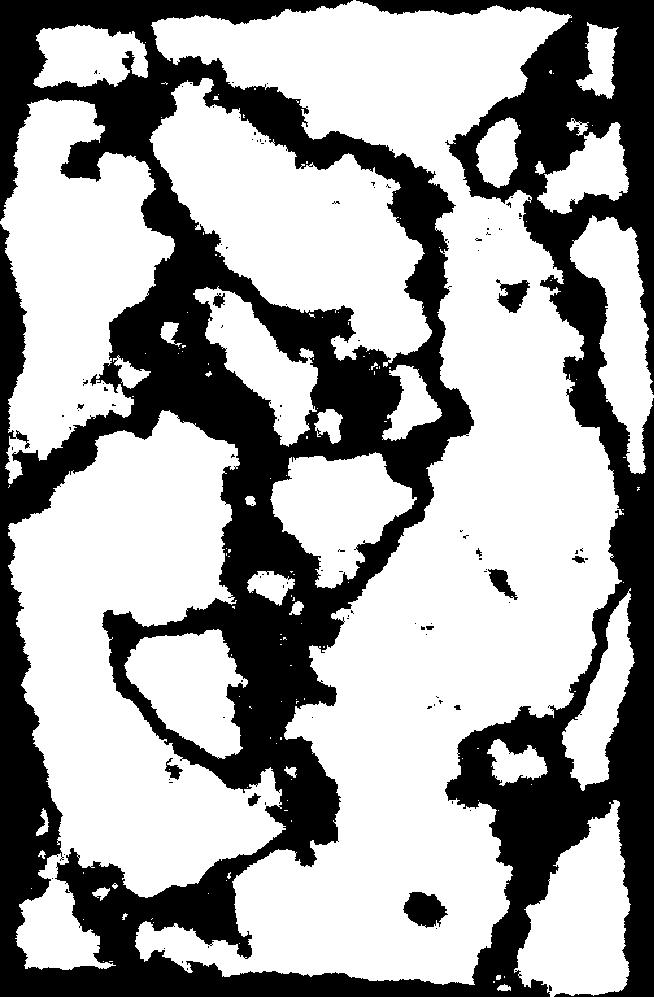}
  \caption{Margin = 6}
\end{subfigure}%
\begin{subfigure}{.2\textwidth}
  \centering
  \includegraphics[height=4cm]{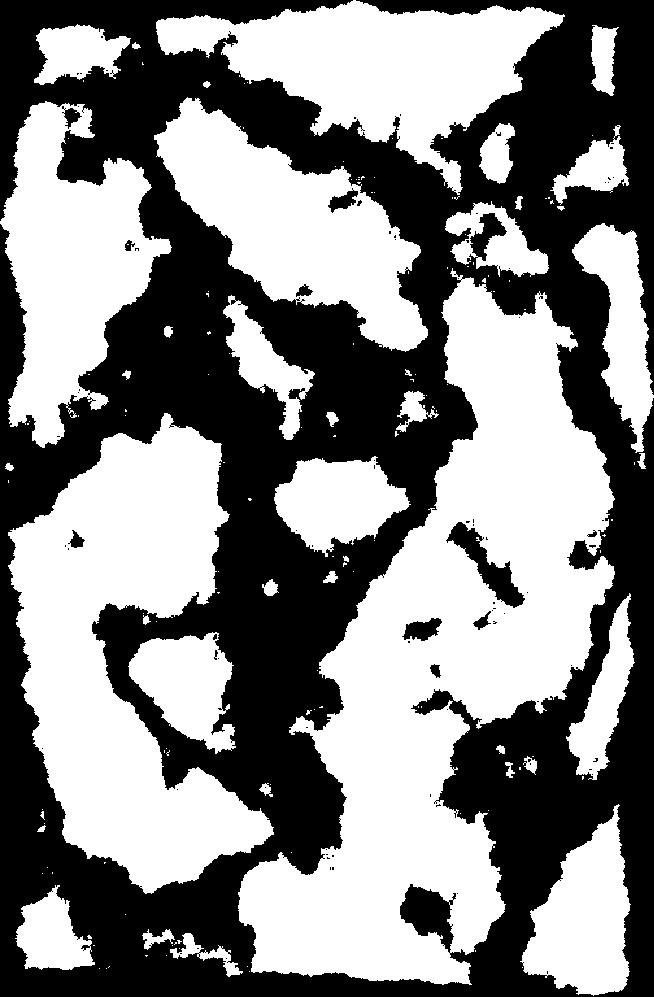}
  \caption{Margin = 9}
\end{subfigure}%
\caption{Impact of the margin on the creation of fragments. We used a margin of 6 for the test data.}
\label{fig:frag-margin}
\end{figure*}

The average value $\bar{v}$ of the diamond-square image is computed.
We binarize the image by selecting as foreground all pixels whose values $v$ are such that $\left|v-\bar{v}\right|\geq m$, where $m$ is a manually-defined margin.
The effect of $m$ is the increase of the margin between fragments.
This is illustrated in \cref{fig:frag-margin}.
While for small values of $m$, it would be possible to match fragments using only their shapes, larger values allow the use of the shape only as hint.

In order to avoid straight edges at the sides of the images, we use a one-dimensional version of the diamond-square to produce non-linear borders on the four sides of the image.
These edges are computed in the same way as the cuts of the rectangular fragments, which are described below.

\subsubsection{Rectangular Fragments}
The rectangular fragments are produced by vertical and horizontal non-linear cuts of the text area.

The three parameters for producing a cut are the approximate position $p$, the roughness $r$ and the amplitude $a$.
Then, it is computed in the following way.
An array of length $L$ is created, and its first and last values are set randomly in $\left[-a, a\right[$.
Then, we set the middle value to a random value in $\left[-r\cdot a, r\cdot a\right[$.
The three defined values split the array in two halves~--~we repeat recursively this process on each half, with $r\cdot a$ instead of $a$, until the whole array is filled.
Then, we add $p$ to each value of the array.
For example, if we want to cut vertically an image in two ``roughly equal'' parts, we set $L$ to the height of the image, and $p$ to half of its width.

The width of the cut, which is the amount of pixels that are erased on both sides of the center of the cut, is computed in the same way, however with a smaller initial $a$, and an $r$ closer to 1.
This makes the width of the stroke more stable, and the borders of the fragment slightly less chaotic.

\begin{figure*}
\centering
\begin{subfigure}{.25\textwidth}
  \centering
  \includegraphics[height=4cm]{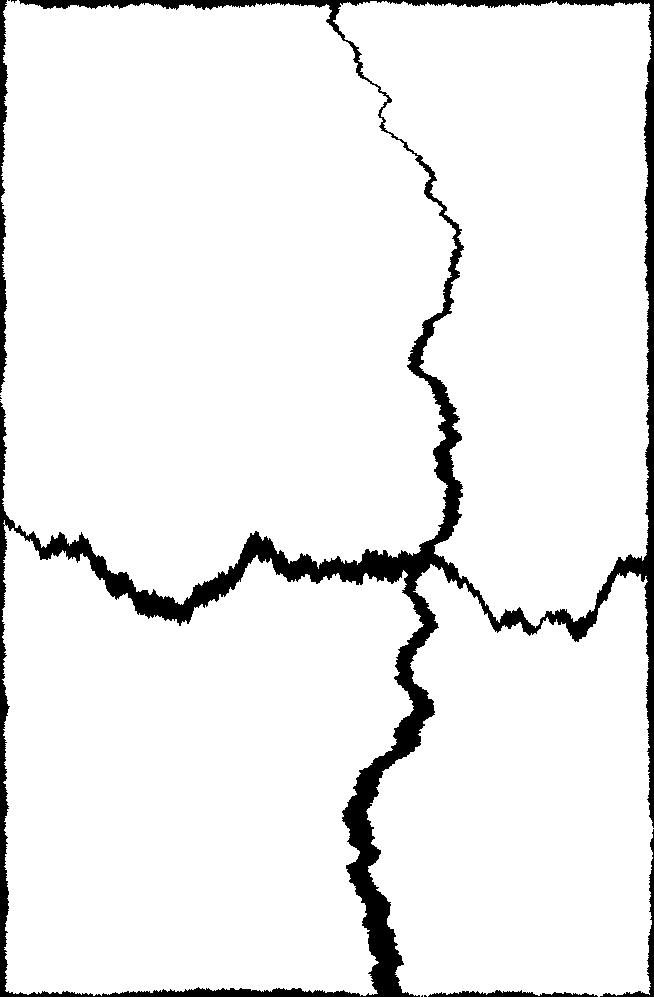}
\end{subfigure}%
\begin{subfigure}{.25\textwidth}
  \centering
  \includegraphics[height=4cm]{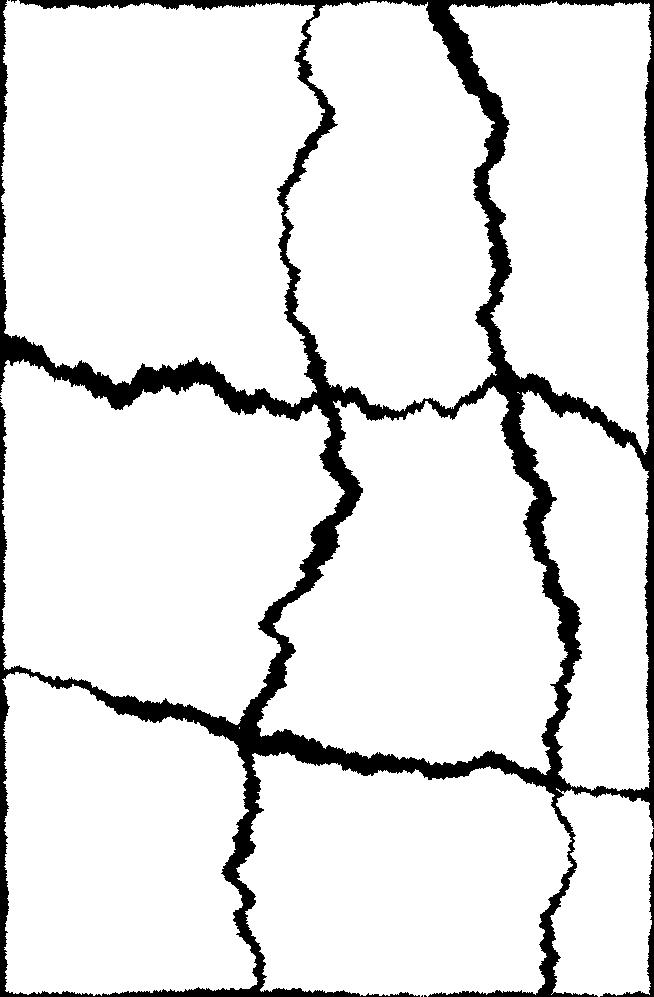}
\end{subfigure}%
\begin{subfigure}{.25\textwidth}
  \centering
  \includegraphics[height=4cm]{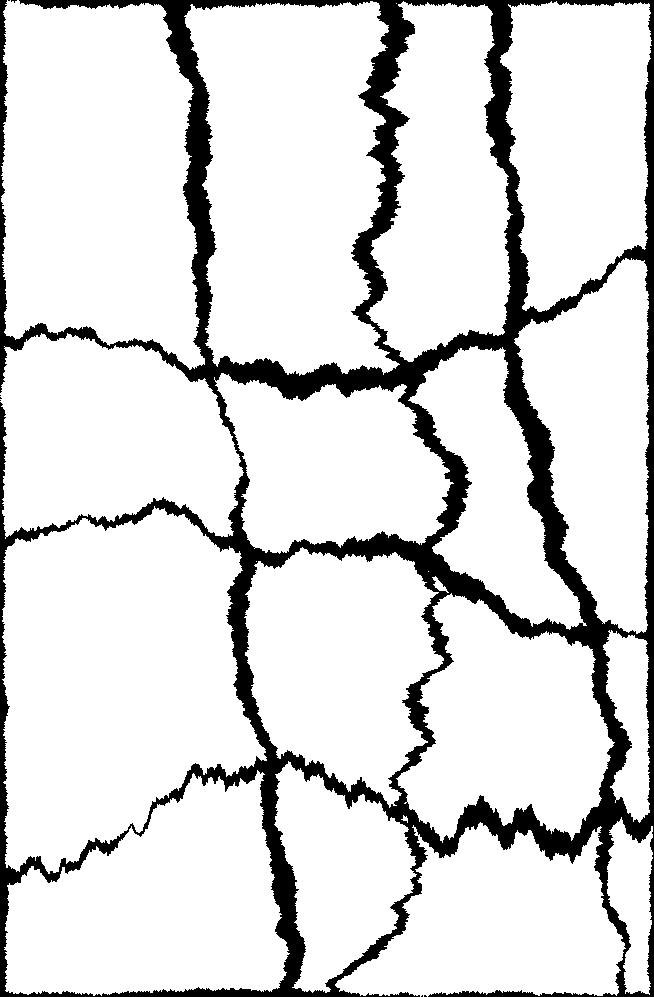}
\end{subfigure}%
\begin{subfigure}{.25\textwidth}
  \centering
  \includegraphics[height=4cm]{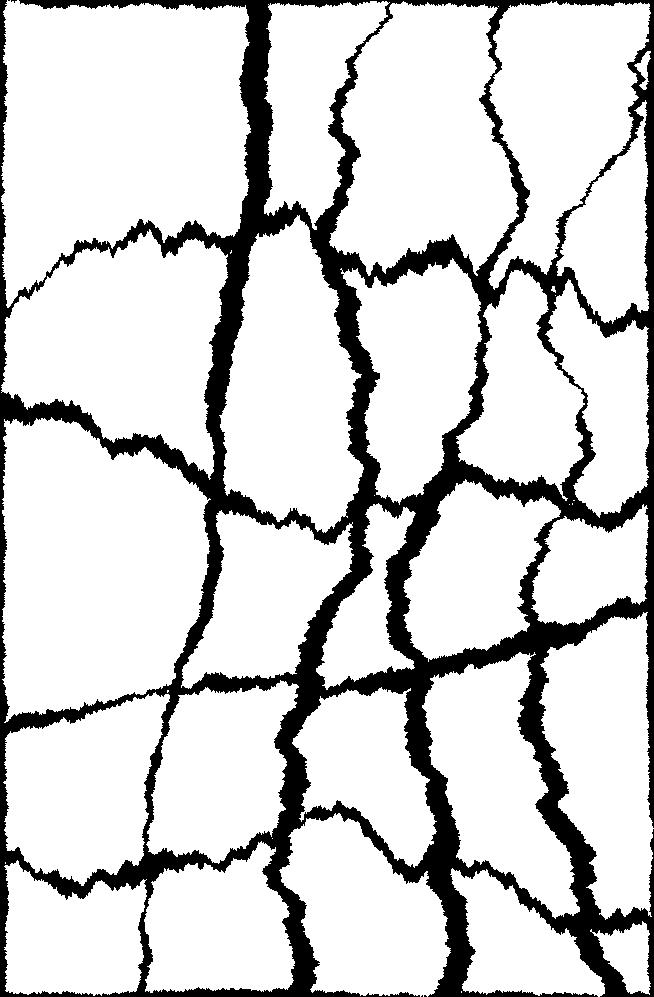}
\end{subfigure}
\caption{Illustration of rectangular fragments generation. Using the same parameters, the number of cuts can be adjusted to produce the desired amount of fragments per text area.}
\label{fig:frag-nxn}
\end{figure*}

An arbitrary number of horizontal and vertical cuts can then be obtained, as shown in \cref{fig:frag-nxn}.
The variance of the cuts width makes a perfect fragment matching using only borders impossible, and thus the task more challenging.
However, shape information still provides hints, as the borders are still correlated.

\subsection{Dataset Origin and Dataset Split}

\paragraph{Origin}
For the training set, we used images of the publicly available Historical-IR19 test set\footnote{\url{https://zenodo.org/record/3262372}}~\cite{Christlein19Comp}. 
This consists of data from different sources, and of different nature: manuscripts, charters, and letters.

For the test set, we used new data, \ie not used in any previous image retrieval competition, consisting of manuscripts and letters. 
The manuscripts are a corpus of books written in the European Middle Ages, mostly from the \nth{9} to the \nth{15} century CE.
These manuscripts are mainly anonymous because writers seldomly signed their work during this period.
We selected sequences of consecutive pages in homogeneous parts of the manuscripts, making the assumption that a single person wrote them.
A paleographer %
checked these sequences to ensure there is no change of writing style within them.

The letters have been provided by the University Library Basel.\footnote{\url{https://www.unibas.ch/}}
Their writers are given in metadata, and we discarded some samples due to too little text or quality issues, such as very faint ink or strong bleed-through.

To ensure that the style of the handwriting can be used for retrieving fragments, all of them should contain some text.
This was not done in the same way for the training set and the test set.

For the training set, the central image part (\SI{25}{\percent} of their surface) has been fragmented.
Then, thumbnails of the fragments have been inspected visually, and the ones without visible text have been removed.
Images have been re-fragmented with the second method when not enough fragments passed the check, and this process was executed until a sufficient number of fragments was obtained.

As this process was time-consuming, another approach has been used for the test set.
We labeled with bounding boxes the main text area of each page, so that fragments of a sufficient size necessarily contain some text.
To ensure an even distribution of text over the labeled area, we have omitted some parts of the main text from the bounding box if their inclusion would lead to text gaps.
We first fragmented images using the first method.
Then, in order to reach \num{20000} fragments, we iteratively deleted the fragments from images which had produced the lowest amount of them, and re-fragmented them with the second method.
The amount of rectangular fragments thus produced for each image, as depicted in \cref{fig:frag-nxn}, was chosen depending on the surface of the text bounding box.

\paragraph{Dataset Split}
The training set constitutes \num{101706} fragments written by \num{8717} writers generated from \num{17222} document images. 
Note, that we have not forbidden the use of additional training data, such as the CLaMM datasets~\cite{Cloppet16,Cloppet17}, the HistoricalWI'17 dataset~\cite{Fiel17ICDAR} or the Historical-IR19 dataset~\cite{Christlein19Comp}. 
Yet, we believe that the provided number of fragments are enough to enable the use of deep neural networks commonly used in image classification. 

The test set contains \num{20019} fragments of \num{1152} writers generated from \num{2732} images.
There is no overlap between the training and test sets, which means that no writer from one set is present in the other one.

\section{Methods}\label{sec:methods}

\subsection{Baseline}
In total, we obtained five submissions from four participants. Additionally, we evaluated the method of \etal{Nicolaou}~\cite{Nicolaou15ICDAR} as our baseline. 
A special variation of Local Binary Patterns (LBP), designed for document image analysis, are computed for radii 1 to 12.
The patterns of each radius are globally pooled into a histogram of 256 bins and normalized.
The histograms are then concatenated, they embed the texture of each text image into a 3072-dimensional vector.
The embedded samples are then mapped to $\mathbb{R}^{200}$ along the principal components of the test-set representing a retrieval scenario where we have access to all embeddings of the test set.
As a final step the Hellinger kernel is applied on the vectors and afterwards normalized by their $\ell^2$ norm. 
The distances are computed by the Manhattan distance. 
For more details refer to~\cite{Nicolaou15ICDAR}.
The exact implementation employed is available online.\footnote{\url{https://github.com/anguelos/wi19_evaluate/tree/master/srslbp}}

\subsection{University of Groningen}
Sheng He and Lambert Schomaker from the University of Groningen submitted a method based on the FragNet~\cite{He20}.
The FragNet was specifically created for writer identification using word snippets. 
It has two pathways: a feature pyramid pathway, which is used to extract feature maps, and a fragment pathway, which is trained to predict the writer based on fragments extracted from the input image and the feature maps of the feature pyramid.
The fragments are used as follows. 
Possible black borders coming from the fragment creation process are removed by computing horizontal/vertical projections.
Each fragment is tiled in non-overlapping blocks of size $64\times128$ because FragNet expects inputs of dimension $64\times128$.
For this task, only cross-entropy loss is used, instead of triplets loss. 
Furthermore, only 250 writers that have more than few hundreds samples were used for training. 
Please note that only 30 epochs instead of 200 were computed due to a hardware issue.
For inference, all blocks of a fragment are passed through the FragNet and the logits are averaged resulting in a 512-dimensional feature vector for each fragment. 
The distances are computed using Euclidean distance.

\subsection{Université de Technologie de Belfort-Montbéliard}
Abderrazak Chahi, Youssef El merabet, Yassine Ruichek, and Raja Touahni from the Laboratoire CIAD UMR 7533, Université de Technologie de Belfort-Montbéliard submitted the following method. 
For robust and reliable feature extraction, the team exploited deep convolutional neural networks (CNNs) as a powerful approach to characterize the writing style of historical handwritten fragments. 
Therefore, two ResNet50 models are fine-tuned using different input data: Net1 uses the entire fragment image while for Net2, the input image is divided into four subblocks, \ie one vertical and one horizontal split. 
Hence, Net2 is using about \num{400000} training samples in total.
For training, the Adam optimizer is used with a learning rate of 0.0001, 10 epochs and a mini-batch size of 40.
For inference, the activations of both networks are concatenated. 
More specifically, the activations of the average pooling layer consisting of a 2048-deep feature vector of each network are concatenated resulting into a 4096 global descriptor for each fragment.
Finally, the distance computation is performed using $\chi^2$ distance. 
The team submitted two variants of their method. 
The first one (model$_\text{writer}$) is trained with writer labels and the second one (model$_\text{page}$) with page labels. 

\subsection{University of Bourgogne Franche-Comte} %
Michel Chammas, Abdallah Makhoul and Jacques Demerjian from the Femto-ST Research Institute at the University of Bourgogne Franche-Comte submitted the following method.
It is based on the approach of Christlein et al.~\cite{Christlein17ICDAR} with some improvements. First, SIFT was used to detect keypoints and extract initial descriptors. PCA was used to reduce the dimensionality of the descriptors from 128 to 32. Afterwards, the SIFT descriptors are clustered using $k$-means. Then, patches extracted from the same SIFT keypoints were used to train a ResNet20 CNN using the corresponding cluster ids of the SIFT descriptors as targets. The CNN activation features of the penultimate layer were encoded using a multi-VLAD approach and normalized to generate a global descriptor vector. Then, an Incremental PCA~\cite{Goel19} with whitening was applied to reduce the dimensionality of the feature vector. At the end, the cosine distance is computed to produce the results as similarity measures.

\subsection{University of Tébessa}
Abdeljalil Gattal and Chawki Djeddi from the University of Tébessa submitted the following method. 
Two different configurations of oriented Basic Image Features (oBIFs) columns histograms ($\sigma={2,4}$ and $\sigma={1,8}$)~\cite{Abdeljalil16,Abdeljalil18} are extracted from the grayscale fragment samples and concatenated for generating a feature vector. 
The oBIF parameter $\epsilon$ is fixed to $0.01$.
Distance computation is carried out using correlation distance.
The technique does not require any further preprocessing and the features are directly extracted from the full writing samples, \ie no binarization is needed.

\begin{table*}[t]
\centering
        \caption{Competition results \subref{tab:writer} per writer and \subref{tab:page} per image. All values given in percent.}
        \label{tab:results}
\subcaptionbox{Task 1 (writer)\label{tab:writer}}{
    \begin{tabular}{llcccc}
    \toprule
    & Method &  \map  & Accuracy & Pr@10 &Pr@100\\
    \midrule
  Baseline
    & SRS-LBP  &
33.4    &60.0     &46.8   &45.9\\
    \midrule
Groningen & FragNet & 
\phantom{0}6.4    &32.5   &16.8   &14.5\\
\multirow{2}{*}{Belfort} & TwoPath$_\text{writer}$ & 
33.5    &\textbf{77.1}   &\textbf{53.1}   &\textbf{50.4}\\
& TwoPath$_\text{page}$ & 
25.2    &61.1   &41.2   &44.1\\
Bourgogne & ResNet20$_\text{ssl}$ & 
\textbf{33.7}    &68.9   &52.5   &46.5\\
    Tebessa & oBIF & 
    24.1    &55.4   &39.2   &37.9\\
    \bottomrule
        \end{tabular}
}\,
\subcaptionbox{Task 2 (image)\label{tab:page}}{
    \begin{tabular}{llcccc}
    \toprule
    & Method &  \map & Accuracy  & Pr@10 &Pr@100\\
    \midrule
    Baseline
    & SRS-LBP &
18.5    &25.7   &25.9   &53.1
    \\
    \midrule
Groningen & FragNet & 
\phantom{0}4.1    & \phantom{0}8.4   &\phantom{0}6.9   &16.6\\
\multirow{2}{*}{Belfort} & TwoPath$_\text{writer}$ & 
\textbf{22.6}    &\textbf{36.4}   &\textbf{31.2}   &\textbf{58.9}\\
& TwoPath$_\text{page}$ & 
17.4    &27.4   &24.6   &52.6\\
Bourgogne & ResNet20$_\text{ssl}$ & 
18.4    &24.1   &26.2   &53.2\\
Tebessa & oBIF & 
16.1    &23.0     &22.4   &45.5\\
    \bottomrule
    \end{tabular}
}
\end{table*}
\section{Evaluation}
\label{sec:evaluation}

\subsection{Error Metrics}
\label{sct:error-metrics}
Each participant was required to hand in a $20019\times20019$ distance matrix. We evaluated the results of the participants in a leave-one-image-out cross-validation manner. This means that every image of the test set will be used as query for which the remaining test images are ranked. The metrics are then averaged over all queries.

Our test-set is unbalanced class-wise. Queries can have from four relevant fragments per writer up-to 69 samples to be retrieved.
Different metrics are evaluated and reported.
The metric that defines the winner is mean Average Precision (mAP)~\cite{nicolaou2018non} computed from the participants' distance matrices.
Mean Average Precision is computed as follows. For each query, the average precision (AP) is computed. Therefore, the precision over all ranks $r$ is computed. Given the retrieved list of size $S$ in which $R$ documents are relevant for the query $q$, then the AP is computed as:
\begin{equation}
    \mathrm{AP}_q = \frac{1}{R} \sum_{r=1}^S \mathrm{Pr}_q (r) \cdot \text{rel}_q(r) \;,
\end{equation}
where $\mathrm{Pr}_q(r)$ is the precision at rank $r$ and $\text{rel}_q(r)$ is an indication function returning $1$ if the document at rank $r$ is relevant and $0$ otherwise. 
The mAP also approximates the area under the curve of the precision recall curve~\cite{larson2010introduction}, \ie the curve that is created when plotting the precision as a function of recall. The evaluation system, which computes the mAP was made available to the participants.\footnote{\url{https://github.com/anguelos/wi19_evaluate}}

Furthermore, we report the Top-1 accuracy, \ie the average precision at rank $1$.
We also provide Pr@10 and Pr@100 as approximations of the usability of systems by human experts representing what a digital humanities expert could go through respectively when investigating quickly and when investigating thoroughly.
Pr@10 and Pr@100 are calculated as 
\begin{equation}
\label{eq:precison}
    \mathrm{Pr@K} = \frac{1}{N}\sum_{n=1}^N {\frac{R_{n,K}}{min(R_{n,N}, K)}} \;,
\end{equation}
where $R_{n,i}$ is the number of relevant samples for query $n$ up to rank $i$, $N$ is the number of samples. The $min(,K)$ in the denominator was added so that a perfect method can achieve a perfect score for every query sample.

\begin{figure*}[t]
    \centering
    \subcaptionbox{\label{fig:size}}{
			\includegraphics[width=0.45\linewidth,height=0.26\textheight]{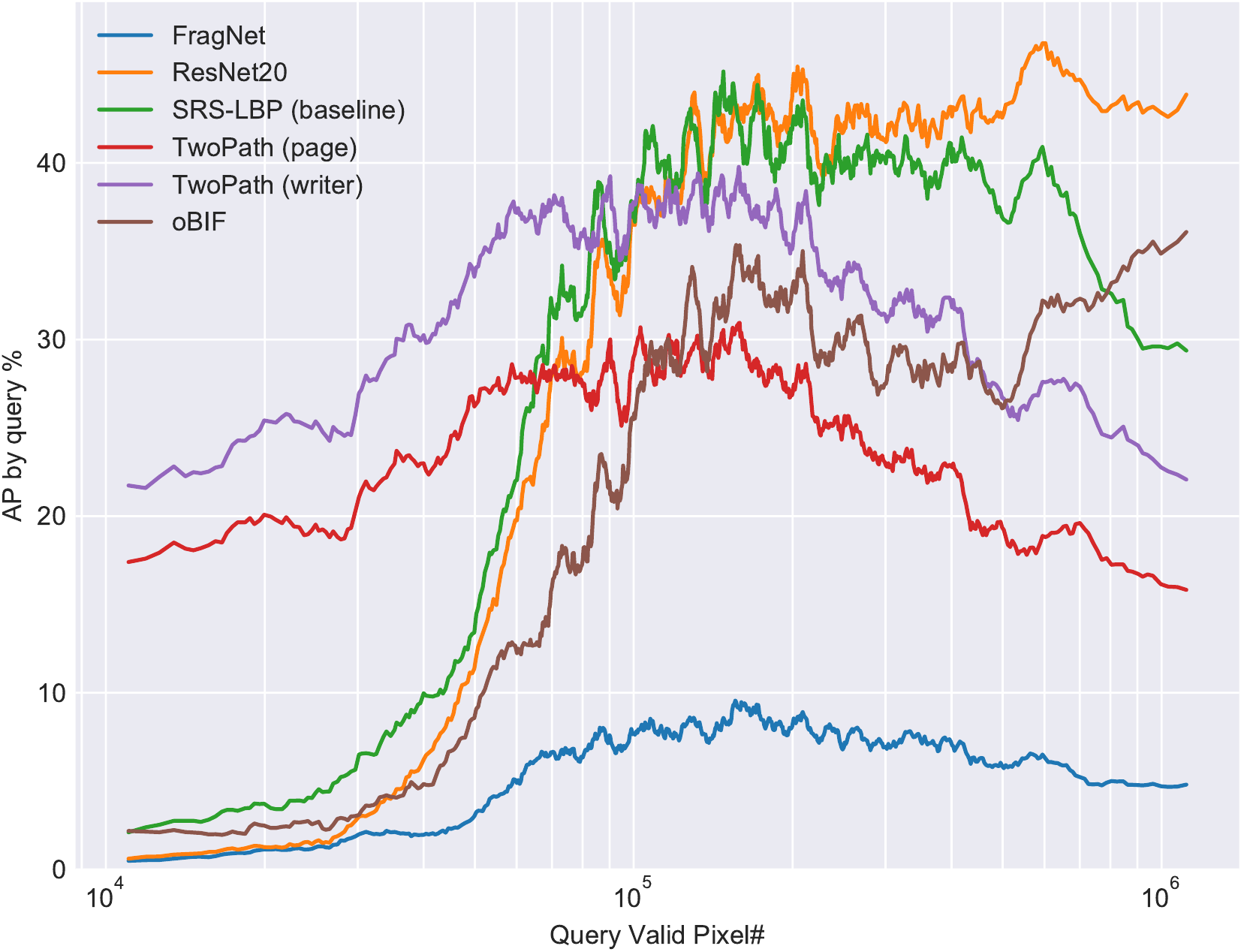}
    }\quad
    \subcaptionbox{\label{fig:bar}}{
			\includegraphics[width=0.45\linewidth,height=0.26\textheight]{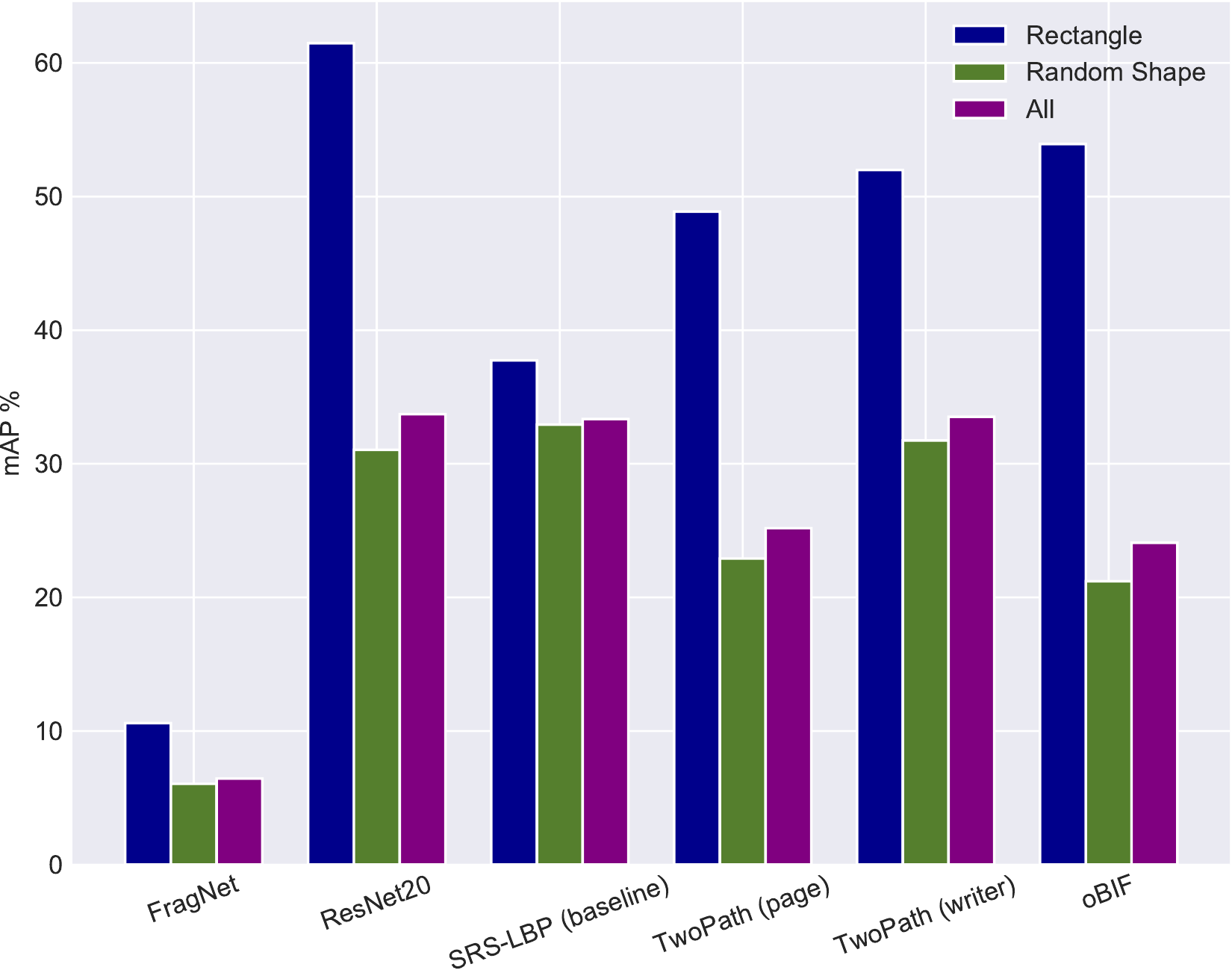}
    }
    \caption{\subref{fig:size} Average precision (for writer labels) versus number of valid pixels used in each query sample, plots are smoothed by 500 samples (a form of local \map), \subref{fig:bar} shows the difference in average precision between random-shaped and rectangular-shaped fragments.}
    \label{fig:pxl_ap}
\end{figure*}
\subsection{Results}

We first evaluate the full test dataset in two different modalities: (a) per writer %
and (b) per page, %
where we interpret page as the full image, \ie if it is a book scan containing two pages it is regarded as one image.
In practice the two modalities are alternative labelings of the test-set.

For the retrieval of fragments written by the same writer (task 1), the results are shown in \cref{tab:writer}. 
The best submission in terms of \map was \emph{Bourgogne}. 
However, the runner-up method (\emph{Belfort}) is in front in terms of accuracy, Pr@10, Pr@100.
For the retrieval of fragments of the same image (task 2), \cf \cref{tab:page}, the method \emph{Belfort} is superior in all metrics. 
Interestingly, the model trained with writer labels (TwoPath$_\text{writer}$) outperforms the other one (TwoPath$_\text{page}$) in both tasks, although the latter was specifically trained for the image task.
An advantage of the ResNet20$_\text{ssl}$ in contrast to the other deep learning-based methods is that it does not need any writer information. 
The \emph{Tebessa} method does not train any neural network and ranks behind the baseline method.
A possible reason might be the influence of the background, which is automatically down-weighted in the baseline method. 
The baseline method performs well and its \map is close to the two leading methods. 
The \emph{Groningen} method result does not perform well for fragment retrieval. 
Maybe the quite irregular shapes with different amount of text caused problems for this method. 
Please further note that this method was designed for word images and had some training issues.

\subsection{Secondary Results} %

First, we evaluate the behavior of the methods in regards to the number of valid pixels. 
In \cref{fig:frag-examples}, valid pixels are all non-black pixels including foreground (ink) and background (paper).
\cref{fig:size} shows plots of the average precision when varying the number of queries in respect to their pixels. 
We notice an interesting behavior of the different methods. In the low pixel regime (\SIrange{1e4}{1e5}{px}), the TwoPath models are superior. However, when the number of pixels increases, the ResNet20$_\text{SSL}$ method takes over, closely followed by the baseline method.
A possible explanation could be the different receptive field size. If the fragments are small, a deeper model (ResNet50 in the TwoPath model) might be more beneficial than the ResNet20$_\text{SSL}$ method, which needs a certain number of local descriptors and benefits from this in the encoding stage. Also the number of SIFT keypoints and thus used patches on the fragment edges is higher in smaller fragments

\cref{fig:bar} shows the mAP each method achieves with respect to the fragment generation algorithm used to generate each query sample.
Samples generated with the rectangle algorithm, are only \SI{8.7}{\percent} of the total test-set so the impact they have in the overall mAP is quite weak.
It should also be pointed out that the rectangle-based fragments have a much higher density of valid pixels (0.60) in comparison to the random-based ones (density: 0.35). 
When using rectangular-shaped samples, the results are much higher for most methods. 

\section{Discussion}
It has long been a matter of speculation whether writer identification systems, specially neural network based methods,  encapsulate specific handwriting attributes or generic texture properties that happen to work well for writer identification. The two alternative writer and page labelings provide insights to this question, and it seems this depends on the system. We hypothesise that the page task is more grounded to general texture (paper, ink-color, etc.) while the writer task is more grounded to higher order features probably associated with what is speculated to be the soft-biometric properties of handwriting. %
Another finding worth pointing out is that, according to \cref{fig:size}, there is no best overall method; depending on the image size, different methods clearly take the lead. This indicates that these methods could successfully be fused with not too much effort.
It specifically seems that ResNet20$_\text{SSL}$ is superior to the other methods by far for the near-rectangular samples, which might be caused by the different sampling procedure.

\section{Conclusion}\label{sec:conclusion}
This competition targeted the retrieval of document fragments. 
Therefore, a large training and test dataset for fragment retrieval was created using the diamond-square algorithm.
In total, four teams participated. Three of the submitted methods are based on deep CNNs. 
In the writer-based evaluation, the winning entry uses a method based on self-supervised learning of the descriptors. 
Another method fuses the outputs of two neural networks, which were trained using different inputs. 
This method is superior in the task of finding the correct fragments of a page and also achieves a better Top-1 accuracy in the first task.
In a fragment size study, we show that the methods differ greatly in their accuracy in respect to the number of valid pixels in the query samples.

\textbf{Acknowledgement}: 
This work has been partly supported by the Cross-border Cooperation Program Czech Republic – Free State of Bavaria ETS Objective 2014–2020 (project no.\ 211).

\bibliographystyle{IEEEtran}
{\small
\bibliography{references}
}
\end{document}